\documentclass[man, 12pt, a4paper]{article}
\usepackage{arxiv}
\usepackage[utf8]{inputenc} 
\usepackage[T1]{fontenc}    
\usepackage{hyperref}       
\usepackage{url}            
\usepackage{booktabs}       
\usepackage{amsfonts}       
\usepackage{nicefrac}       
\usepackage{microtype}      
\usepackage{lipsum}
\usepackage{graphicx}
\usepackage[utf8]{inputenc}
\usepackage{setspace} 
\usepackage{natbib}
\usepackage{dirtytalk}
\usepackage{etoolbox}
\usepackage{wrapfig}
\usepackage{siunitx, booktabs}
\usepackage{mwe}
\usepackage{graphicx}
\usepackage{listings}
\usepackage[capposition=top]{floatrow}
\usepackage{amsmath}
\usepackage{hyperref}
\usepackage{algorithm}
\usepackage[noend]{algpseudocode}
\usepackage{multirow}
\usepackage{tikz}
\usepackage{pgfplots}
\usetikzlibrary{patterns}
\usepackage{csvsimple}
\usepackage{pgfplotstable}
\pgfplotsset{every tick label/.append style={font=\tiny}}
\pgfplotsset{
	box plot width/.initial=4em,
	box plot/.style={
		/pgfplots/.cd,
		black,
		only marks,
		mark=-,
		mark size=\pgfkeysvalueof{/pgfplots/box plot width},
		/pgfplots/error bars/.cd,
		y dir=plus,
		y explicit,
	},
	box plot box/.style={
		/pgfplots/error bars/draw error bar/.code 2 args={%
			\draw [line width=0.20mm]  ##1 -- ++(\pgfkeysvalueof{/pgfplots/box plot width},0pt) |- ##2 -- ++(-\pgfkeysvalueof{/pgfplots/box plot width},0pt) |- ##1 -- cycle;
		},
		/pgfplots/table/.cd,
		y index=2,
		y error expr={\thisrowno{3}-\thisrowno{2}},
		/pgfplots/box plot
	},
	box plot top whisker/.style={
		/pgfplots/error bars/draw error bar/.code 2 args={%
			\pgfkeysgetvalue{/pgfplots/error bars/error mark}%
			{\pgfplotserrorbarsmark}%
			\pgfkeysgetvalue{/pgfplots/error bars/error mark options}%
			{\pgfplotserrorbarsmarkopts}%
			\path ##1 -- ##2;
		},
		/pgfplots/table/.cd,
		y index=4,
		y error expr={\thisrowno{2}-\thisrowno{4}},
		/pgfplots/box plot
	},
	box plot bottom whisker/.style={
		/pgfplots/error bars/draw error bar/.code 2 args={%
			\pgfkeysgetvalue{/pgfplots/error bars/error mark}%
			{\pgfplotserrorbarsmark}%
			\pgfkeysgetvalue{/pgfplots/error bars/error mark options}%
			{\pgfplotserrorbarsmarkopts}%
			\path ##1 -- ##2;
		},
		/pgfplots/table/.cd,
		y index=5,
		y error expr={\thisrowno{3}-\thisrowno{5}},
		/pgfplots/box plot
	},
	box plot median/.style={
		/pgfplots/box plot
	}
}
\pgfplotsset{yticklabel style={text width=1.0em,align=right}}
\pgfplotsset{compat=1.12}
\definecolor{olivegreen}{RGB}{85,107,47}
\definecolor{orange}{RGB}{229,148,0}
\definecolor{marine}{RGB}{0,32,96}
\definecolor{maroon}{RGB}{178, 50, 50}
\definecolor{atomictangerine}{rgb}{1.0, 0.6, 0.4}
\definecolor{bittersweet}{rgb}{1.0, 0.44, 0.37}
\definecolor{bondiblue}{rgb}{0.0, 0.58, 0.71}
\definecolor{darkcoral}{rgb}{0.8, 0.36, 0.27}
\definecolor{darkcyan}{rgb}{0.0, 0.55, 0.55}
\definecolor{capri}{rgb}{0.0, 0.75, 1.0}
\definecolor{emerald}{rgb}{0.31, 0.78, 0.47}
\definecolor{carnationpink}{rgb}{1.0, 0.65, 0.79}
\definecolor{inchworm}{rgb}{0.7, 0.93, 0.36}
\usepackage{mathtools}
\usepackage{amsmath,amssymb,amsfonts}
\usepackage{multirow}
\usepackage{graphicx}
\usepackage{amsbsy}
\usepackage{pdflscape}
\usepackage{booktabs}
\usepackage{xcolor}
\usepackage{colortbl}
\usepackage[caption=false]{subfig}
\usepackage{algorithm}
\usepackage[noend]{algpseudocode}
\usepackage{multirow}

\title{Evolutionary Multi-Objective Optimisation for Fairness-Aware Self Adjusting Memory Classifiers in Data Streams}

\author {
Pivithuru Thejan Amarasinghe\\
La Trobe University\\
Melbourne\\
Australia\\
\texttt{p.amarasinghe@latrobe.edu.au}
\And
Diem Pham\\
La Trobe University\\
Melbourne\\
Australia\\
Can Tho University\\
Can Tho\\
Vietnam\\
\texttt{ThiXuanDiem.PHAM@latrobe.edu.au}
\And
Binh Tran\\
La Trobe University\\
Melbourne\\
Australia\\
\texttt{B.Tran@latrobe.edu.au}
\And
Su Nguyen\\
RMIT University\\
Melbourne\\
Australia\\
\texttt{su.nguyen@rmit.edu.au}
\And
\;\;\;\;\;\;\;\;\;\;\;\;\;\;\;\;\;\;\;\;\;\;\;\;Yuan Sun\\
\;\;\;\;\;\;\;\;\;\;\;\;\;\;\;\;\;\;\;\;\;\;\;\;La Trobe University\\
\;\;\;\;\;\;\;\;\;\;\;\;\;\;\;\;\;\;\;\;\;\;\;\;Melbourne\\
\;\;\;\;\;\;\;\;\;\;\;\;\;\;\;\;\;\;\;\;\;\;\;\;Australia\\
\texttt{\;\;\;\;\;\;\;\;\;\;\;\;\;\;\;\;\;\;\;\;\;\;Yuan.Sun@latrobe.edu.au}
\And
\;\;\;\;\;\;\;\;\;\;Damminda Alahakoon\\
\;\;\;\;\;\;\;\;\;\;La Trobe University\\
\;\;\;\;\;\;\;\;\;\;Melbourne\\
\;\;\;\;\;\;\;\;\;\;Australia\\
\texttt{d.alahakoon@latrobe.edu.au}
}

\begin{document}
\maketitle
\begin{abstract}
    This paper introduces a novel approach, evolutionary multi-objective optimisation for fairness-aware self-adjusting memory classifiers, designed to enhance fairness in machine learning algorithms applied to data stream classification. With the growing concern over discrimination in algorithmic decision-making, particularly in dynamic data stream environments, there is a need for methods that ensure fair treatment of individuals across sensitive attributes like race or gender. The proposed approach addresses this challenge by integrating the strengths of the self-adjusting memory K-Nearest-Neighbour algorithm with evolutionary multi-objective optimisation. This combination allows the new approach to efficiently manage concept drift in streaming data and leverage the flexibility of evolutionary multi-objective optimisation to maximise accuracy and minimise discrimination simultaneously. We demonstrate the effectiveness of the proposed approach through extensive experiments on various datasets, comparing its performance against several baseline methods in terms of accuracy and fairness metrics. Our results show that the proposed approach maintains competitive accuracy and significantly reduces discrimination, highlighting its potential as a robust solution for fairness-aware data stream classification. Further analyses also confirm the effectiveness of the strategies to trigger evolutionary multi-objective optimisation and adapt classifiers in the proposed approach.
\end{abstract}


\section{Introduction}
\label{introduction}
In recent years, machine learning (ML) algorithms have been applied to support decision-making across various domains such as finance and criminal \cite{ghadami2022data,jena2020machine,patil2022system}. However, a significant concern within these applications is the potential for discrimination that affects individuals with respect to factors such as race, gender, and socioeconomic status. For instance, healthcare providers and health plans that rely on clinical algorithms for decision-making may inadvertently discriminate against certain patients, resulting in some groups losing access to healthcare services \cite{cary2023mitigating,huang2022evaluation}. Racial bias in ML poses a significant challenge, and addressing this issue is essential to ensure fairness in implementing ML \cite{mehrabi2021survey}.

Fairness-aware ML has gained significant attention in addressing the issue of discrimination within algorithms. Since discrimination often arises from sensitive features such as gender or race, techniques such as removing sensitive attributes, applying feature selection, or assigning weights to features to mitigate discrimination have played a crucial role in mitigating discrimination. However, the trade-off arises because trying to reduce discrimination can unintentionally lead to less accurate models. Therefore, minimising discrimination and maximising accuracy can be considered as conflicting objectives.

Swarm intelligence (SI), inspired by animal behaviours, is a subset of evolutionary computation \cite{mavrovouniotis2017survey}. SI algorithms leverage swarm interactions for adaptation and cooperation, making them promising for learning in dynamic environments and addressing challenges in changing contexts. \cite{pevska2019swarm}. SI algorithms such as Particle Swarm Optimisation \cite{ali2020particle} and Salp chain-based optimisation \cite{al2020salp} have been applied in feature weighting methods, allocating higher weights to informative features and vice versa \cite{nino2021feature}. SI has also demonstrated significant potential in multi-objective optimisation tasks \cite{zhou2011multiobjective}. In recent work, SMPSO \cite{nebro2009smpso} has shown promise in improving fairness for online classification, FOMOS \cite{pham2024fairness}.

Discrimination issues can occur in many ML applications, especially in situations where data is constantly changing, like data stream scenarios \cite{pessach2022review}. While fairness in ML has received attention, there are limited studies on fairness in streaming data. Several fairness-aware classification algorithms, such as FAHT~\cite{zhang2019faht}, 2FAHT~\cite{zhang2020flexible}, and FAS Stream~\cite{pham2022fairness}, have been proposed for learning streaming data. These methods aim to minimise the impact on sensitive features during model learning. However, they require predefined constants for feature weights (FAHT and 2FAHT) or proportions for optimising feature weights (FAS Stream).

To overcome the limitations above, this paper proposes a new algorithm called Evolutionary Multi-Objective Optimisation for fairness-aware Self Adjusting Memory Classifiers (EMOSAM). The proposed algorithm combines the strength of SAMKNN \cite{losing2016knn} in managing concept drift and the flexibility of EMO \cite{zhou2011multiobjective,nebro2009smpso} in optimising feature weights and handling multiple conflicting objectives. Specifically, SAMKNN is responsible for streaming data classification, and SMPSO optimises feature weights that SAMKNN use to mitigate discrimination and maximise accuracy. The main contributions of this paper are as follows.

\begin{itemize}
    \item A new incremental classification method called EMOSAM to learn fairness-aware SAMKNN classifiers for data stream classification.
    \item An EMO component within EMOSAM to determine feature weights that maximise accuracy and minimise discrimination simultaneously for SAMKNN.
    \item Strategies to trigger EMO and adapt SAMKNN classifiers by selecting non-dominated solutions obtained by EMO. 
\end{itemize}

The remainder of the paper is organised as follows. The background and related works are presented in Section~\ref{sec:background}. The details of our method are introduced in Section~\ref{sec:method}. Section~\ref{sec:experimentandresult} shows the dataset and experimental setups for validating our method. The conclusions and future work are discussed in Section \ref{sec:conclusion}.

\section{Background and Related Works}
\label{sec:background}
\subsection{Data Stream Classification}
There have been many algorithms for the classification of data streams proposed in the literature \cite{mehta2017concept}. Tree-based algorithms such as the Hoeffding Tree Classifier (HT) \cite{hulten2001mining}, the Hoeffding Adaptive Tree Classifier (HAT)~\citep{bifet2007learning}, and the Adaptive Random Forest Classification~\cite{gomes2017adaptive} can quickly adapt decision tree models to incorporate new concepts in streaming data~\cite{gomes2019machine}. Meanwhile, the Nearest Neighbour approach, such as k-Nearest-Neighbour (KNN), is also widely used for learning from data streams and performs consistently well across benchmark datasets. Ensemble-based algorithms synthesise the results of individual learners to make final decisions, exploiting the adaptability of multiple learners to handle diverse data distributions~\cite{krawczyk2017ensemble}. Studies showcasing the accuracy weighted ensemble classifier \cite{wang2003mining}, the hierarchical structure of the ensemble \cite{yin2015de2,you2016simple}, and dynamic ensemble sizes \cite{xu2017dynamic} underscore the benefits of this approach. In particular, SAMKNN \cite{losing2016knn} stands out as a state-of-the-art method that deals with changes in data streaming using both short-term memory for current concepts and long-term memory for past concepts.

\subsection{Fairness-aware Machine Learning}
\subsubsection*{Discrimination Measures} 
The initial phase to improve fairness within learnt models involves assessing discrimination in ML models. Several discrimination measures, including statistical parity, equalised odds, and absolute between-ROC area~\cite{vzliobaite2017measuring}, have been introduced for this purpose. Statistical parity, widely used in numerous studies~\cite{zhang2019faht,le2022survey,iosifidis2019fairness}, calculates the percentage difference between the unprotected and protected groups when assigned a positive class ($Y=1$). In this paper, we adopt statistical parity as the discrimination measure due to its simple implementation and popular applications in the fairness-aware ML literature.

Consider a dataset $D$ used for binary classification tasks. Each instance in $D$ consists of a set of features denoted by $x$, and a corresponding class label denoted by $Y$, which can take on values of either $0$ or $1$. Additionally, we have a sensitive feature denoted by $S$, which can take on values from the set $\{u, p\}$. Here, $u$ represents the unprotected group (e.g., females) and $p$ represents the protected group (e.g., males). The predicted values $\hat{Y}$ are obtained from ML models. The discrimination of a given ML model, based on the statistical parity criterion, can be calculated as follows.
\begin{equation}
\label{eq:disc}
    Disc(D) = P(\hat{Y} = 1 \mid S = p) - P(\hat{Y} = 1 \mid S = u),
\end{equation}
where $P(\hat{Y} = 1 \mid S = p)$ represents the probability of the protected group receiving a positive classification, while $P(\hat{Y} = 1 \mid S = u)$ denotes the probability of the unprotected group receiving a positive classification. As a result, the range of $Disc(D)$ is between -1 and 1. A value of 0 corresponds to no discrimination, a value of 1 indicates discrimination against the protected group, and a value of -1 signifies discrimination against the unprotected group. This equation can also be used to assess discrimination in the original dataset $D$ by replacing $\hat{Y}$ with $Y$.
\subsubsection*{Fairness-aware ML for Streaming Data}
Fairness-aware ML techniques aim to mitigate discrimination while maintaining the accuracy of their predictions. These techniques predominantly employ three approaches: pre-processing, in-processing, and post-processing, which correspond to intervening before, during, or after the model training process to enhance fairness, respectively.

Pre-processing techniques aim to mitigate discrimination in a dataset. Two popular methods within this approach are ``massaging'' and ``reweighting''~\cite{kamiran2009classifying,calders2009building}. The former involves relabelling data, while the latter assigns weights to different sensitive groups. In streaming classification, fairness enhancing interventions (FEI) adhere to a pre-processing approach, modifying data chunks before inputting them into a classification model \cite{iosifidis2019fairness}. These data modification methods aim to mitigate discrimination. 

FABBOO (Fairness-Aware Batch Based Online Oversampling) \cite{iosifidis2020online} is also a pre-processing technique tailored for fairness in online learning with class imbalance. It aims to address biases by oversampling the minority class. However, this method can lead to increased computation time. Additionally, FABBOO relies on predefined thresholds for resampling, potentially affecting its effectiveness.

In-processing methods adjust models by embedding fairness measurements into their objective functions to address discrimination. These techniques focus on mitigating the impact of sensitive features within ML models. In the decision tree family, the adaptive fairness-aware decision tree classifier (FAHT) \cite{zhang2019faht}, 2FAHT \cite{zhang2020flexible}, and fairness-enhancing and concept-adapting decision tree classifier (FEAT) \cite{zhang2020feat} are derived from Hoeffding Tree (HT). These models introduce novel splitting criteria that consider discrimination during tree growth, showcasing success in mitigating bias. However, they exhibit a trade-off between accuracy and discrimination and rely on user input. The FAS stream \cite{pham2022fairness} handles discrimination using swarm intelligence for feature weighting, but the balance between accuracy and discrimination is user-defined.

Post-processing methods alter the decision boundary of a model or adjust its prediction labels. Most of the methods are examined in static scenarios. Some techniques introduce additional prediction thresholds, as demonstrated in \cite{hardt2016equality}, to address discrimination, while others, such as in \cite{calders2010three}, modify the decision boundary of models for fairness. These methods primarily target the output of a classifier. Adapting these techniques to online learning poses challenges, as the evolving data distribution in streaming data may lead to changes in the boundary or predictions.

\subsection{EMO in Fairness Machine Learning}
Fairness ML aims to generate models that are fair and accurate. However, previous studies~\cite{liu2022accuracy,zhang2022mitigating} have shown that fairness and accuracy are often conflicting objectives, meaning that improving one usually comes at the expense of the other.
Therefore, multi-objective optimisation, especially evolutionary multi-objective optimisation (EMO), has been applied to address these conflicts~\cite{padh2021addressing,liu2022accuracy,yu2022towards}.
For example, \citet{rehman2022fair} proposed an EMO algorithm to select features that enhance both fairness and accuracy objectives.\citet{zhang2022mitigating,zhang2021fairer} introduce a learning framework that uses EMO to simultaneously optimise various metrics, including accuracy and multiple fairness measures, within ML models. The results indicate the ability of the EMO algorithms to discover a more diverse set of fair models using solutions on the Pareto front to develop ensemble models with good performance. However, these methods have been predominantly applied to a limited set of benchmark datasets in static scenarios. In data stream learning, FOMOS \cite{pham2024fairness} employs EMO with HT, automatically optimising feature weights for fairer and more accurate trees. However, FOMOS uses a singular strategy for leader selection from Pareto solutions and triggers EMO solely when classifier discrimination increases. This motivates our paper to introduce strategies for selecting solutions from the Pareto front produced by EMO specifically tailored for streaming data. 

\newcommand{\pushline}{\Indp}
\clearpage
\section{Our Method}
\label{sec:method}

\begin{figure}[!hp]
    \centering
    \includegraphics[width=0.48\textwidth]{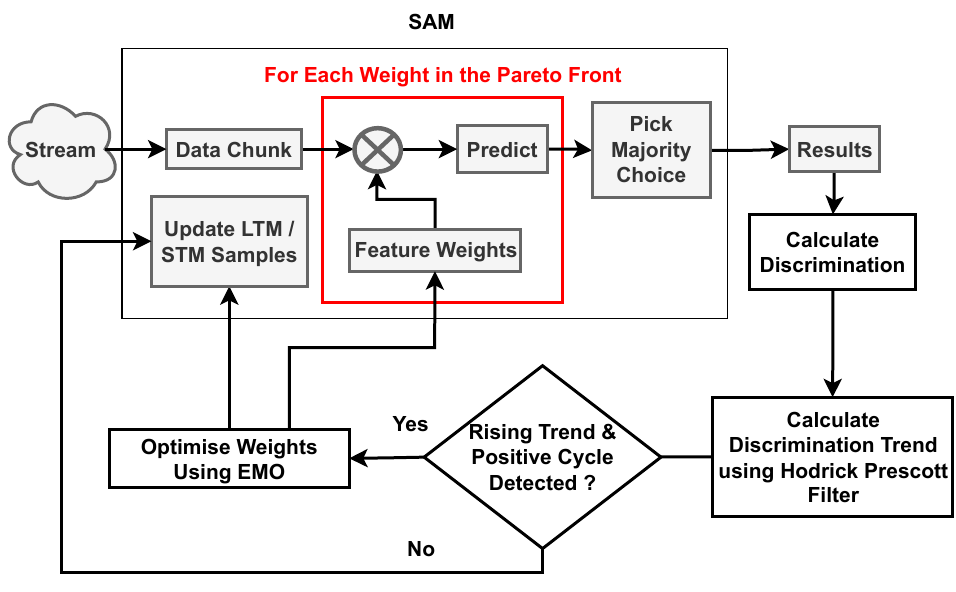}
    \caption{An overview of our proposed EMOSAM algorithm for data stream classification.}
    \label{fig:solution_overview}
\end{figure}

We introduce a novel approach called EMOSAM, which combines EMO with the SAM algorithm to improve data stream classification. Our approach takes into account both model accuracy and fairness (or discrimination) as objectives. Figure~\ref{fig:solution_overview} provides an overview of the EMOSAM approach. Similar to ensemble methods, our algorithm maintains a set of weak learners that have different trade-offs between model accuracy and fairness. These weak learners have different feature weights that are evolved using an EMO algorithm. If a significant model discrimination is detected, there is a good chance that the previously identified feature weights are no longer suitable for the new data. In such cases, the EMO algorithm is used to evolve a new set of feature weights based on the discrimination trend. The remainder of this section will provide details for each key component of EMOSAM.
\vspace{-0.5cm}
\vspace{-0.5cm}
\subsection{Weighted SAMKNN for Classification}

SAMKNN, proposed by \citet{losing2016knn}, is a cutting-edge approach for classifying data streams. It uses the self-adjusting memory (SAM) architecture, which is specifically designed to handle heterogeneous concept drift in streaming data. SAMKNN uses long-term and short-term memory to make decisions when faced with changes in streaming data. Short-term memory (STM) represents the current concept, while long-term memory (LTM) preserves established knowledge as long as it aligns with the STM. Incoming data points are stored in the STM, and a cleaning process ensures that the LTM remains consistent with the STM. When the size of the STM is reduced, its discarded knowledge is transferred to the LTM. The accumulated knowledge is compressed when the available space is exhausted. During prediction, both the STM and LTM are taken into account, based on their past performance.

SAMKNN is known for its ability to handle concept drift, a common issue in streaming data classification. However, one drawback of SAMKNN is that it uses all features for predictions, which can make it susceptible to bias risks. In order to address this limitation, our proposed method aims to enhance SAMKNN by introducing feature weights. These feature weights are applied to each feature to control its influence on the distance calculation for STM and LTM in SAMKNN. The specific steps for making predictions using SAMKNN with feature weights are outlined in Algorithm~\ref{alg:sampredict}.

\begin{algorithm}[!tb]
  \caption{SAMKNN Prediction with Feature Weights}
  \label{alg:sampredict}
\begin{algorithmic}
  \State {\bfseries Input:} data points $X$, feature weights $\alpha$, SAMKNN classifier $\mathcal{M}$.
  \State {\bfseries Output:} the predicted class lable $\hat{y}$ for each data point $\mathbf{x} \in X$.
  \State Transform the data points $X$, the STM samples and LTM samples in $\mathcal{M}$ by multiplying the feature weights $\alpha$. 
  \For{each data point $\mathbf{x} \in X$}
  \State $N_1 \leftarrow$ the $k$ nearest neighbours of $\mathbf{x}$ in the STM samples. 
  \State $N_2 \leftarrow$ the $k$ nearest neighbours of $\mathbf{x}$ in the LTM samples. 
  \State $N_3 \leftarrow$ the $k$ nearest neighbours of $\mathbf{x}$ in the STM and LTM. 
  \State $y_1 \leftarrow$ the majority class among $N_1$. 
  \State $y_2 \leftarrow$ the majority class among $N_2$.
  \State $y_3 \leftarrow$ the majority class among $N_3$.
  \State $\hat{y} \leftarrow$ selected from $y_1$, $y_2$, and $y_3$ based on past accuracy. 
  \EndFor
\end{algorithmic}
\end{algorithm}

The original SAMKNN classifier can be seen as a specific case of the weighted SAMKNN, where all feature weights are set to one. Determining the appropriate feature weights can potentially enhance the accuracy and fairness of SAMKNN. However, optimising these weights is challenging. In our approach, the feature weights are initialised randomly and refined using an EMO algorithm when the current classifier fails to mitigate discrimination.

\subsection{Trigger for Feature Weight Optimisation}

Applying EMO to optimise feature weights can be expensive, so it is not necessary to apply it every time a new window arrives. Instead, we need to determine the appropriate time to re-optimise the feature weights based on discrimination. In the EMOSAM approach, we maintain a list of discrimination values for previous windows. Within each new window, we analyse the discrimination list ($PD$ in Algorithms~\ref{alg:emosam_algorithm}) to determine if there is an increasing trend in discrimination. We treat the model discriminations for past windows as time series data and use the HP filter~\cite{hodrick1997postwar} to identify any trends in discrimination. The HP filter decomposes a time series ($y_t$) into a trend component ($\Gamma_t$), a cyclical component ($C_t$), and an error component ($\epsilon_t$): 
\begin{equation}
\label{eq:hp-filter}
   y_t = \Gamma_t + C_t + \epsilon_t. 
\end{equation}
If a rising trend and a positive cycle are detected, we apply EMO to search for new feature weights and reset the discrimination list. Furthermore, having a positive cyclic component for the last value of the past discrimination values will determine whether discrimination is starting to reduce or not. HP filter is applied because of its robustness against randomness and noise, which is the case in data stream classification. Validation of the HP filter's robustness will be further examined in the ablation study (Section~\ref{sec:comparewithother}).

\subsection{SMPSO for Optimising Feature Weights}
\label{sec:method}

We address the issue of determining feature weights for SAMKNN as a multi-objective optimisation problem. To optimise the feature weights, we employ the Speed-constrained Multi-objective PSO (SMPSO) algorithm \cite{nebro2009smpso}, which aims to balance two conflicting objectives: (1) maximising accuracy and (2) minimising discrimination. We choose SMPSO because it is widely recognised as a highly effective multi-objective PSO algorithm, particularly in dynamic optimisation problems \cite{mavrovouniotis2017survey}. The success of SMPSO in approximating Pareto fronts and its faster convergence towards these fronts, as demonstrated in previous work \cite{nebro2009smpso}, make it a suitable choice for streaming learning tasks that require rapid convergence.

SMPSO starts by creating a swarm of particles that represent potential solutions, as well as an archive of non-dominated solutions within the swarm (i.e., solutions that are not dominated by any other solutions in terms of all the objectives). In each iteration, the positions and velocities of the particles are updated, and a polynomial mutation technique \cite{deb2011multi} is used to generate new solutions. The resulting particles are then evaluated to update both the swarm and the archive. After a specified number of iterations, SMPSO returns the non-dominated solutions found in the archive.

Each solution or particle generated by SMPSO is a feature weight vector ($\alpha$), which will be used to calculate the distance between data points in the SAMKNN classifier. The optimisation process aims to find the optimal $\alpha$ that balances accuracy and discrimination. For a given number of features $d$, a particle $\alpha$ can be represented as a $d$-dimensional vector $\alpha = [\alpha_{1}, \alpha_{2}, \alpha_{3}, \dots, \alpha_{d}]$, where $\alpha_{i}$ is a real number between 0 and 1. Additionally, a particle maintains an objective value vector $obj = [obj_{acc}, obj_{disc}]$, where $obj_{acc}$ and $ obj_{disc}$ represent the accuracy and discrimination achieved by applying $\alpha$ to the SAMKNN model as feature weights. The discrimination $ obj_{disc}$ is the absolute statistical parity in Equation (\ref{eq:disc}).

Algorithm~\ref{algo:evaluation} shows the process of evaluating a particle to determine its objective values. The input to this algorithm includes the current data chunk $D_t$, the position of a particle $\alpha$, and the SAMKNN classifier. In order to assess the effectiveness of each particle, the feature weights are incorporated into SAMKNN to generate predictions for $D_t$. The resulting objective values, denoted as $obj = [obj_{acc}, obj_{disc}]$, represent the outcome of the evaluation.

\begin{algorithm}[!hp]
  \caption{Multi-objective evaluation}
  \label{algo:evaluation}
\begin{algorithmic}
  \State {\bfseries Input:} Training data $D_t = \{(\mathbf{x}_1, y_1), \dots, (\mathbf{x}_n, y_n)\}$, SAMKNN classifier $\mathcal{M}$, solution position or feature weights $\alpha$.
  \State {\bfseries Output:} objective values $obj = [obj_{acc}, obj_{disc}]$.
  \State Predict for $D_t$ using $\mathcal{M}$ with feature weights $\alpha$ (Algorithm~\ref{alg:sampredict}). 
  \State $obj_{acc}$ $\leftarrow$ calculate accuracy of the prediction.
  \State $obj_{disc}$ $\leftarrow$ calculate discrimination of the prediction.
\end{algorithmic}
\end{algorithm}
\vspace{-0.5cm}

\subsection{Feature Weight Selection for SAMKNN}
The selection of feature weight vectors from the Pareto front generated by SMPSO is a challenging task, as each vector represents a unique trade-off between model accuracy and fairness. In the EMOSAM approach, instead of choosing a single weight vector, we consider all the weight vectors on the Pareto front. Specifically, we utilise each weight vector to transform the features and input them into the SAMKNN algorithm for making predictions. This approach enables us to generate an ensemble of SAMKNNs. To determine the final prediction of our ensemble, we employ a majority voting scheme, selecting the prediction with the highest number of votes. The details of the selection of the feature weight as well as other components of EMOSAM are described in Algorithm~\ref{alg:emosam_algorithm}.

\begin{algorithm}[!tb]
  \caption{EMOSAM Algorithm}
  \label{alg:emosam_algorithm}
\begin{algorithmic}
  \State {\bfseries Input:} Data Stream $\mathcal{D} = \{D_1, \dots, D_T\}$, and trend threshold $\Phi$. 
  \State PF $\leftarrow$ randomly initialised feature weight vectors. 
  \State PD $\leftarrow$ an empty list to store past model discriminations.
  \For{$t$ from $1$ to $T$}
  \State Get the features $X_t$ of the data chunk $t$ from $D_t$.
  \State Initialise a list to store predictions: 
  $L \leftarrow \{\}$.
  \For{each feature weight vector $\alpha \in$ PF}
  \State $\hat{Y}_{\alpha} \leftarrow$ predict for $X_t$ with feature weights $\alpha$ (Algorithm~\ref{alg:sampredict}).
  \State Add the predictions $\hat{Y}_{\alpha}$ to the list $L$.
  \EndFor
  \State $\hat{Y}_t \leftarrow$ majority voting of the predictions in $L$. 
  \State $Disc_t \leftarrow$ compute model discrimination based on $\hat{Y}_t$ and $X_t$. 
  \State Add $Disc_t$ into the list PD and remove $Disc_{t-5}$ if exists.
  \State Compute trend ($\Gamma_t$) and cycle ($C_t$) in PD using HP filter.
  \If{last value of $\Gamma_t$ $\ge \Phi$ {\bf and} last value of $C_t$ $> 0$}
  \State Optimise SAMKNN feature weights using SMPSO on $D_t$.
  \State PF $\leftarrow$ the Pareto front generated by SMPSO.
  \State Remove all elements in PD. \vspace{0.15cm} 
  \EndIf
  Fit SAMKNN with $D_t$ and update STM and LTM samples.
  \EndFor
\end{algorithmic}
\end{algorithm}
\section{Experiment and Result Analysis}
\label{sec:experimentandresult}

\subsection{Datasets and Experiment Design}
\label{sec:datasetanddesign}
Six classification datasets, commonly used in fairness learning \cite{le2022survey}, are used to examine the performance of EMOSAM. Table \ref{tab:datasets} shows the information of each dataset, including the number of instances $(N)$, dimension $(d)$, the sensitive feature $(SF)$, the unprotected group $(UP)$, and the discrimination level $(Disc)$ calculated based on Equation (\ref{eq:disc}).

\begin{table}[!tb]
\centering
\caption{The datasets used in our experiments.}\label{tab:datasets}
\resizebox{0.48\textwidth}{!}{
\begin{tabular}{lrrccr}
\toprule
\textbf{Dataset} & \textbf{N} &  \textbf{d} & \textbf{SF} & \textbf{UP} & \textbf{Disc (\%)}\\
\midrule
Law & 20,798 & 23 & Race & Non\_White & 19.83 \\
Credit & 30,000 & 91 & Gender & Female & 3.39 \\
Adult &  32,561 & 36 & Gender & Female & 19.63 \\
Bank & 41,188 & 53 & Age & Age$\le$25 or Age$\ge$60 & 22.15 \\
Dutch &  60,420 & 11 & Gender & Female & 29.51 \\
Census &  284,556 & 370 & Gender & Female & 7.53 \\
\bottomrule
\end{tabular}
}
\end{table}

To examine the performance of EMOSAM,
EMOSAM is compared with seven baseline methods in terms of accuracy and discrimination. While HT, HAT \citep{bifet2007learning} and SAM represent fairness-unaware methods, FAHT \cite{zhang2019faht} and FABBOO \cite{iosifidis2020online} represent traditional fairness-aware methods, FOMOS~\cite{fomos} and FAS Stream~\cite{pham2022fairness} are evolutionary fairness-aware methods. All methods are run with default parameters in the original studies. For FOMOS, FOMOS\_2obj (maximising accuracy and minimising discrimination) is chosen to compare with EMOSAM. As EMOSAM, FOMOS, and FAS Stream are stochastic algorithms, 30 independent runs were conducted for each method on each dataset with different seeds.

Table \ref{tab:para} shows the experiment's parameter settings. All methods utilise a window size of 1000, a common choice in fairness learning on streaming data. For SMPSO, the population size is set to 30, and the number of iterations is 10 to support online learning. The remaining SMPSO parameters are adopted from the jMetalPy \cite{benitez2019jmetalpy}.

\begin{table}[t]
\centering
\caption {Parameter Settings.}
\begin{tabular}{llr}
\toprule
&\textbf{Parameters} & \textbf{\#Settings}  \\ 
\midrule
    All methods & $window\_size$  & $1000$ \\ 
    \midrule
    EMOSAM & max STM size & $5000$\\
    & Trend threshold $\phi$ & 10\% \\
    \midrule
    SMPSO & Population size &	$30$ \\ 
    & Maximum iterations  &	$10$ \\ 
    & $c_1, c_2$ &	$1.49445$     \\ 
    & $w$ &	$0.1$     \\
    & Leader size & $100$ \\ 
    \bottomrule
\end{tabular} \label{tab:para}
\end{table}

\subsection{Comparisons with Baseline Methods}
\label{sec:comparewithother}

\begin{table}[!tb]
 \footnotesize
 \caption {Accuracy and discrimination obtained by EMOSAM and other methods. Since all the standard deviations of accuracy are smaller than $0.01\%$, we do not display them to save space. Note that FAHT fails on the law and bank datasets.}
 \begin{tabular}{ll|rrc|rrc}
 \toprule
 \multirow {2}{*} {\textbf{Dataset}} & \multirow {2}{*} {\textbf{Method}}
     & \multicolumn{2}{c}{\textbf{Accuracy (\%)}}& & \multicolumn{2}{c}{\textbf{Discrimination (\%)}}\\ 
    \cline{3-8}
    & &\textbf{Best} &\textbf{Avg}& \textbf{T1} & \textbf{Best} &\textbf{Avg $\pm$ Std}& \textbf{T2}\\ 
 \toprule
  \multirow {8}{*} {$\mathsf{Law}$}
                       & HT & 89.29  &  & -- & 23.72  &  & + \\
                       & HAT & 89.18 &  & -- & 21.06 &  & + \\ 
                       & SAM & 88.94 &  & \cellcolor[HTML]{87CEFA}= & 7.22 &  & \cellcolor[HTML]{87CEFA}+ \\ 
                       \cline{2-8}
                       & FABBOO & 86.56 &  & + & 0.99 &  & -- \\ 
                        \cline{2-8}
                       & FOMOS & 89.51 & 89.28  & -- & 18.96 & 19.11 $\pm$ 0.0255 & +\\
                       & FAS Stream & 89.10 & 88.92  & \cellcolor[HTML]{87CEFA}+ & 6.37 & 8.02 $\pm$ 0.0104 & \cellcolor[HTML]{87CEFA}+ \\
                       & EMOSAM & 89.32 & 89.11 & & 2.23 & 7.22 $\pm$ 0.0152 & \\
                       \toprule
                       
  \multirow {8}{*} {$\mathsf{Credit}$}
                       & HT & 79.41  &  & -- & 1.58  &  & + \\
                       & HAT & 81.19 &  & -- & 1.58 &  & + \\ 
                       & SAM & 77.27 &  & \cellcolor[HTML]{87CEFA}= & 0.61 &  & \cellcolor[HTML]{87CEFA}+ \\ 
                       \cline{2-8}
                       & FAHT & 78.43 &  & \cellcolor[HTML]{FFDAB9}-- & 0.16 &  & \cellcolor[HTML]{FFDAB9}-- \\ 
                       & FABBOO & 73.22 &  & + & 0.04 &  & -- \\ 
                       \cline{2-8}
                       & FOMOS & 81.41 & 81.36 & -- & 1.07 & 1.19 $\pm$ 0.0002 & +\\ 
                       & FAS Stream & 77.53 & 77.15  & \cellcolor[HTML]{87CEFA}+ & 0.76 & 1.13 $\pm$ 0.0020 & \cellcolor[HTML]{87CEFA}+ \\
                       & EMOSAM & 77.39 & 77.25  & & 0.37 & 0.54 $\pm$ 0.0006 &\\ 
                       \toprule     
 \multirow {8}{*} {$\mathsf{Adult}$}
                       & HT & 81.82  &  & -- & 20.96  &  & + \\
                       & HAT & 81.68 &  & -- & 20.25 &  & + \\ 
                       & SAM & 78.36 &  & \cellcolor[HTML]{87CEFA}+ & 13.06 &  & \cellcolor[HTML]{87CEFA}+ \\ 
                       \cline{2-8}
                       & FAHT & 79.42 &  & -- & 16.78 &  & + \\ 
                       & FABBOO & 74.76 &  & + & 0.33 &  & -- \\ 
                       \cline{2-8}
                       & FOMOS & 82.22 & 82.01  & -- & 17.77 & 18.23 $\pm$ 0.0019 & +\\ 
                       & FAS Stream & 79.67 & 78.80  & \cellcolor[HTML]{87CEFA}= & 10.91 & 13.42 $\pm$ 0.0146& \cellcolor[HTML]{87CEFA}+ \\
                       & EMOSAM & 78.66 & 78.65  & & 9.44 & 11.97 $\pm$ 0.0089 & \\ 
                       \toprule    
  \multirow {8}{*} {$\mathsf{Bank}$}
                       & HT & 88.20  &  & \cellcolor[HTML]{87CEFA}= & 2.87  &  & \cellcolor[HTML]{87CEFA}+\\
                       & HAT & 88.46 &  & -- & 13.55 &  & + \\ 
                       & SAM & 89.15 &  & -- & 11.93 &  & + \\
                       \cline{2-8}
                       & FABBOO & 87.10 &  & \cellcolor[HTML]{87CEFA}+  & 2.26 &  & \cellcolor[HTML]{87CEFA}+ \\ 
                       \cline{2-8}
                       & FOMOS & 88.14 & 88.08  & \cellcolor[HTML]{87CEFA}= & 2.36 & 2.73 $\pm$ 0.0028 & \cellcolor[HTML]{87CEFA}+\\ 
                       & FAS Stream & 89.51 & 89.09 & -- & 8.60 & 13.04 $\pm$ 0.0257& + \\
                       & EMOSAM & 88.66 & 88.16  & & 0.49 & 1.79 $\pm$ 0.0067 & \\ 
                       \toprule          
  \multirow {8}{*} {$\mathsf{Dutch}$}
                       & HT & 81.46  &  & -- & 35.83  &  & + \\
                       & HAT & 81.92 &  & -- & 31.39 &  & + \\ 
                       & SAM & 76.93 &  & \cellcolor[HTML]{87CEFA}+ & 39.57 &  & \cellcolor[HTML]{87CEFA}+ \\ 
                       \cline{2-8}
                       & FAHT & 73.33 &  &  + & 19.37 &  & -- \\ 
                       & FABBOO & 70.86 &  & + & 0.06 &   & --\\ 
                       \cline{2-8}
                       & FOMOS & 81.87 & 81.55  & -- & 27.35 & 28.28 $\pm$ 0.0172 & + \\ 
                       & FAS Stream & 80.27 & 78.88 & + & 15.27 & 18.64 $\pm$ 0.0273& --\\
                       & EMOSAM & 80.25 & 79.42  & & 22.47 & 26.87 $\pm$ 0.0256 & \\ 
                       \toprule               
  \multirow {8}{*} {$\mathsf{Census}$}
                       & HT & 94.57  &  &  -- & 1.72  &  & +\\
                       & HAT & 94.52 &  & -- & 1.33 &  & +\\ 
                       & SAM & 93.99 &  & + & 0.97 &  & -- \\ 
                       \cline{2-8}
                       & FAHT & 94.28 &  & -- & 3.20 &  & +\\ 
                       & FABBOO & 88.18 &  & + & 0.07 &  & --\\ 
                      \cline{2-8}
                       & FOMOS & 94.55 & 94.54& -- & 1.43 & 1.48 $\pm$ 0.0001 & +\\ 
                       & FAS Stream & 93.94 & 93.88 & + & 0.65 & 0.76 $\pm$ 0.0005 & --\\
                       & EMOSAM & 94.37 & 94.22  & & 0.90 & 1.09 $\pm$ 0.0006 &\\ 

\bottomrule                        
\end{tabular} 
\label{tab:withothermethods}
\end{table}

Table~\ref{tab:withothermethods} presents the accuracy and discrimination in percentage obtained by all the compared methods. For stochastic methods (FOMOS, FAS, and EMOSAM), the best and average results of 30 runs are reported. Wilcoxon statistical test is conducted in each comparison to confirm their significance.
Columns $T1$ and $T2$ show the results of this test comparing EMOSAM accuracy and discrimination with each baseline method, respectively. A `$+$' indicates that the accuracy/discrimination of EMOSAM is significantly better than the method in the corresponding row, and conversely, a `$-$' denotes the opposite. More occurrences of the `$+$' symbols indicate superior performance of EMOSAM. We also highlight $T1$ and $T2$ in blue if they contain $(+ +)$, $(= +)$, or $(+ =)$, which means that EMOSAM dominates the corresponding method. On the other hand, if EMOSAM is dominated by the corresponding method, i.e. $(-  -)$, the cells will be highlighted in orange.

\subsubsection*{EMOSAM versus the fairness-unaware methods}
Compared to HT, HAT, and SAM, EMOSAM's accuracy obtained 3 `$+$', 3 `$=$', and 12 `$-$', while its discrimination result is significantly lower in 17 cases out of the 18 comparisons. In particular, the discrimination of EMOSAM is 16\% and 11\% lower than that of HT on Law and Bank, respectively. It is worth noting that HT, HAT, and SAM concentrate solely on improving the accuracy of the models, while EMOSAM considers both accuracy and fairness. However, EMOSAM still can dominate HT on one dataset and SAM on four datasets, as shown in the blue highlighted rows.

\subsubsection*{EMOSAM versus the traditional fairness-aware methods}
FAHT and FABBOO are popular fairness-aware methods for streaming data. The Wilcoxon test results in Table \ref{tab:withothermethods} indicate that EMOSAM's accuracy consistently outperforms FABBOO across all datasets with 10\% and 6\% higher than FABBOO on Dutch and Census, respectively. However, regarding discrimination, FABBOO wins on all datasets except Bank. This shows a strong preference for discrimination in FABBOO. Mixed results are shown when comparing EMOSAM with FAHT. While EMOSAM is dominated by FAHT on Credit, EMOSAM ran successfully and achieved competitive results on all datasets, including the two highly imbalanced datasets (Law and Bank), where FAHT failed to run (i.e. no FAHT results for the two datasets). Overall, the results show that EMOSAM produces prediction models with a better balance between accuracy and discrimination than FABBOO and FAHT.

\subsubsection*{EMOSAM versus evolutionary fairness-aware methods}
Although FOMOS, FAS Stream, and EMOSAM employ evolutionary optimisation methods for feature weighting to maximise accuracy and minimise discrimination, EMOSAM dominates FAS Stream on three datasets and FOMOS on one while not being dominated by any of these methods. The superior results of EMOSAM confirm the contribution of the proposed multiobjective optimisation approach to fairness-aware algorithms.

In general, the results show that EMOSAM generates fairer classifiers than the fairness-unaware algorithms (HT, HAT, and SAM) while maintaining a better balance between accuracy and discrimination than the fairness-aware methods.

\begin{figure*}[!h]
\centering
\subfloat[Law]{
\begin{tikzpicture}
	\begin{axis} [box plot width=0.20em, height=0.30\textwidth,width=0.38\textwidth, grid style={line width=.1pt, draw=gray!10},major grid style={line width=.2pt,draw=gray!30}, xmajorgrids=true, ymajorgrids=true,  major tick length=0.05cm, minor tick length=0.0cm, legend style={at={(1.0,0.60)},anchor=west,font=\scriptsize,draw=none}]
    \addplot[color=emerald,mark=*,mark size=2.5,only marks] coordinates {(23.72, 10.71)};
    \addplot[color=emerald,mark=halfcircle*,mark size=2.5,only marks] coordinates {(21.06, 10.82)};
    \addplot[color=emerald,mark=square*,mark size=2.5,only marks] coordinates {(7.22, 11.06)};
    \addplot[color=capri,mark=diamond*,mark size=2.5,only marks] coordinates {(0.99, 13.44)};
    \addplot[color=carnationpink,mark=triangle*,mark size=2.5,only marks] coordinates {(19.11, 10.72)};
    \addplot[color=carnationpink,mark=halfsquare*,mark size=2.5,only marks] coordinates {(8.02, 11.08)};
    \addplot[color=darkcoral,mark=pentagon*,mark size=3.5,only marks] coordinates {(7.22, 10.89)};
	\end{axis}
\end{tikzpicture}
}
\subfloat[Credit]{
\begin{tikzpicture}
	\begin{axis} [box plot width=0.20em, height=0.30\textwidth,width=0.38\textwidth, grid style={line width=.1pt, draw=gray!10},major grid style={line width=.2pt,draw=gray!30}, xmajorgrids=true, ymajorgrids=true,  major tick length=0.05cm, minor tick length=0.0cm, legend style={at={(1.0,0.60)},anchor=west,font=\scriptsize,draw=none}]
    \addplot[color=emerald,mark=*,mark size=2.5,only marks] coordinates {(1.58, 20.59)};
    \addplot[color=emerald,mark=halfcircle*,mark size=2.5,only marks] coordinates {(1.58, 18.81)};
    \addplot[color=emerald,mark=square*,mark size=2.5,only marks] coordinates {(0.61, 22.73)};
    \addplot[color=capri,mark=halfdiamond*,mark size=2.5,only marks] coordinates {(0.16, 21.57)};
    \addplot[color=capri,mark=diamond*,mark size=2.5,only marks] coordinates {(0.04, 26.78)};
    \addplot[color=carnationpink,mark=triangle*,mark size=2.5,only marks] coordinates {(1.19, 18.64)};
    \addplot[color=carnationpink,mark=halfsquare*,mark size=2.5,only marks] coordinates {(1.13, 22.85)};
    \addplot[color=darkcoral,mark=pentagon*,mark size=3.5,only marks] coordinates {(0.54, 22.75)};
	\end{axis}
\end{tikzpicture}
}
\subfloat[Adult]{
\begin{tikzpicture}
	\begin{axis} [box plot width=0.20em, height=0.30\textwidth,width=0.38\textwidth, grid style={line width=.1pt, draw=gray!10},major grid style={line width=.2pt,draw=gray!30}, xmajorgrids=true, ymajorgrids=true,  major tick length=0.05cm, minor tick length=0.0cm, legend style={at={(1.0,0.60)},anchor=west,font=\scriptsize,draw=none}]
    \addplot[color=emerald,mark=*,mark size=2.5,only marks] coordinates {(20.96, 18.18)};
    \addplot[color=emerald,mark=halfcircle*,mark size=2.5,only marks] coordinates {(20.25, 18.32)};
    \addplot[color=emerald,mark=square*,mark size=2.5,only marks] coordinates {(13.06, 21.64)};
    \addplot[color=capri,mark=halfdiamond*,mark size=2.5,only marks] coordinates {(16.78, 20.58)};
    \addplot[color=capri,mark=diamond*,mark size=2.5,only marks] coordinates {(0.33, 25.24)};
    \addplot[color=carnationpink,mark=triangle*,mark size=2.5,only marks] coordinates {(18.23, 17.99)};
    \addplot[color=carnationpink,mark=halfsquare*,mark size=2.5,only marks] coordinates {(13.42, 21.2)};
    \addplot[color=darkcoral,mark=pentagon*,mark size=3.5,only marks] coordinates {(11.97, 21.35)};
	\end{axis}
\end{tikzpicture}
}	

\subfloat[Bank]{
\begin{tikzpicture}
	\begin{axis} [box plot width=0.20em, height=0.30\textwidth,width=0.38\textwidth, grid style={line width=.1pt, draw=gray!10},major grid style={line width=.2pt,draw=gray!30}, xmajorgrids=true, ymajorgrids=true,  major tick length=0.05cm, minor tick length=0.0cm, legend style={at={(1.0,0.60)},anchor=west,font=\scriptsize,draw=none}]
    \addplot[color=emerald,mark=*,mark size=2.5,only marks] coordinates {(2.87, 11.8)};
    \addplot[color=emerald,mark=halfcircle*,mark size=2.5,only marks] coordinates {(13.55, 11.54)};
    \addplot[color=emerald,mark=square*,mark size=2.5,only marks] coordinates {(11.93, 10.85)};
    \addplot[color=capri,mark=diamond*,mark size=2.5,only marks] coordinates {(2.26, 12.9)};
    \addplot[color=carnationpink,mark=triangle*,mark size=2.5,only marks] coordinates {(2.73, 11.86)};
    \addplot[color=carnationpink,mark=halfsquare*,mark size=2.5,only marks] coordinates {(13.04, 10.91)};
    \addplot[color=darkcoral,mark=pentagon*,mark size=3.5,only marks] coordinates {(1.79, 11.94)};
	\end{axis}
\end{tikzpicture}
}
\subfloat[Dutch]{
\begin{tikzpicture}
	\begin{axis} [box plot width=0.20em, height=0.30\textwidth,width=0.38\textwidth, grid style={line width=.1pt, draw=gray!10},major grid style={line width=.2pt,draw=gray!30}, xmajorgrids=true, ymajorgrids=true,  major tick length=0.05cm, minor tick length=0.0cm, legend style={at={(1.0,0.60)},anchor=west,font=\scriptsize,draw=none}]
    \addplot[color=emerald,mark=*,mark size=2.5,only marks] coordinates {(35.83, 18.54)};
    \addplot[color=emerald,mark=halfcircle*,mark size=2.5,only marks] coordinates {(31.39, 18.08)};
    \addplot[color=emerald,mark=square*,mark size=2.5,only marks] coordinates {(39.57, 23.07)};
    \addplot[color=capri,mark=halfdiamond*,mark size=2.5,only marks] coordinates {(19.37, 26.67)};
    \addplot[color=capri,mark=diamond*,mark size=2.5,only marks] coordinates {(0.06, 29.14)};
    \addplot[color=carnationpink,mark=triangle*,mark size=2.5,only marks] coordinates {(28.28, 18.45)};
    \addplot[color=carnationpink,mark=halfsquare*,mark size=2.5,only marks] coordinates {(18.64, 21.12)};
    \addplot[color=darkcoral,mark=pentagon*,mark size=3.5,only marks] coordinates {(26.87, 20.58)};
	\end{axis}
\end{tikzpicture}
}	
\subfloat[Census]{
\begin{tikzpicture}
	\begin{axis} [box plot width=0.20em, height=0.30\textwidth,width=0.38\textwidth, grid style={line width=.1pt, draw=gray!10},major grid style={line width=.2pt,draw=gray!30}, xmajorgrids=true, ymajorgrids=true,  major tick length=0.05cm, minor tick length=0.0cm, legend style={at={(0.65,0.65)},anchor=west,font=\scriptsize,draw=none}]
    \addplot[color=emerald,mark=*,mark size=2.5,only marks] coordinates {(1.72, 5.43)};\addlegendentry{\tiny HT}
    \addplot[color=emerald,mark=halfcircle*,mark size=2.5,only marks] coordinates {(1.33, 5.48)};\addlegendentry{\tiny HAT}
    \addplot[color=emerald,mark=square*,mark size=2.5,only marks] coordinates {(0.97, 6.01)};\addlegendentry{\tiny SAM}
    \addplot[color=capri,mark=halfdiamond*,mark size=2.5,only marks] coordinates {(3.2, 5.72)};\addlegendentry{\tiny FAHT}
    \addplot[color=capri,mark=diamond*,mark size=2.5,only marks] coordinates {(0.07, 11.82)};\addlegendentry{\tiny FABBOO}
    \addplot[color=carnationpink,mark=triangle*,mark size=2.5,only marks] coordinates {(1.48, 5.46)};\addlegendentry{\tiny FOMOS}
    \addplot[color=carnationpink,mark=halfsquare*,mark size=2.5,only marks] coordinates {(0.76,	6.12)};\addlegendentry{\tiny FAS Stream}
    \addplot[color=darkcoral,mark=pentagon*,mark size=3.5,only marks] coordinates {(1.09, 5.78)};\addlegendentry{\tiny EMOSAM}
	\end{axis}
\end{tikzpicture}
}

\caption{The discrimination ($x$-axis) and error ($y$-axis) in percentage produced by various methods on the test datasets. Our EMOSAM approach is located on the Pareto front for five datasets and is only dominated by FAHT on the Credit dataset.}
\label{fig:scatter}
\end{figure*}
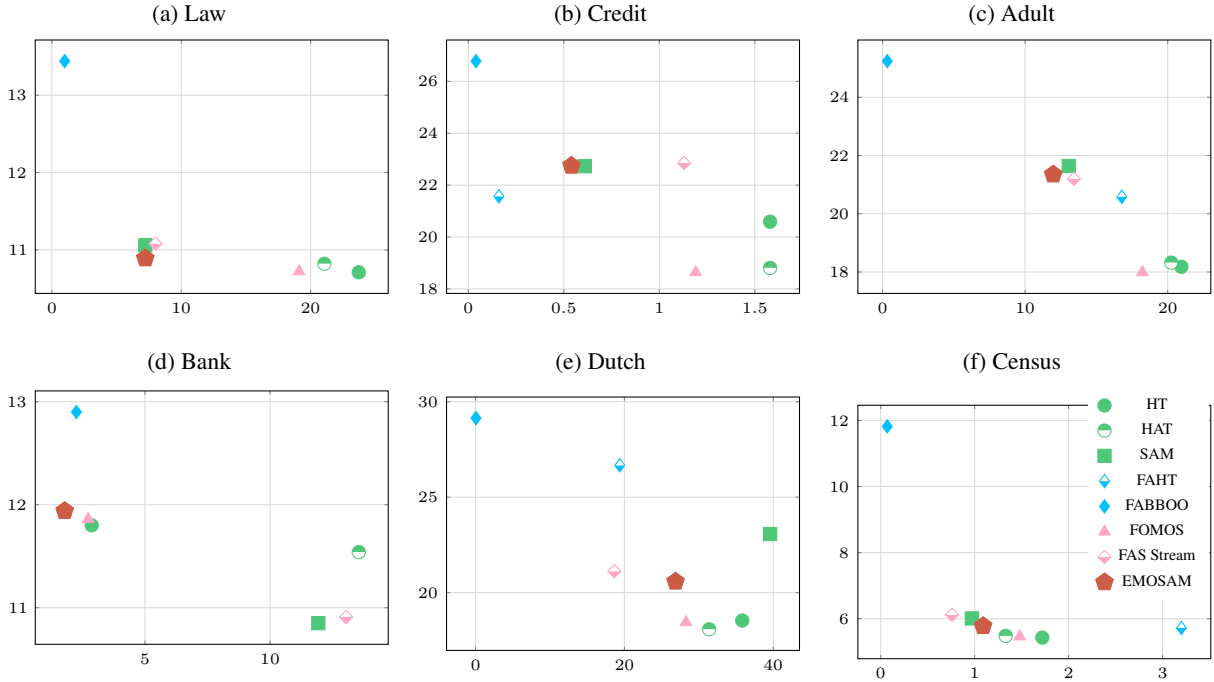

\subsubsection*{Pareto dominance} 
To visualise the dominancy of EMOSAM over the baseline methods, we use scatter plots (Figure \ref{fig:scatter}) with model error on the $y$-axis and discrimination on the $x$-axis. As FOMOS, FAS Stream, and EMOSAM are stochastic algorithms, the average accuracy and discrimination results of the 30 independent runs on each dataset are displayed in the figure. A desirable method exhibits low model error and low discrimination, positioning methods in the bottom left corners as favourable. Three different colours are used to highlight the results of each group of baseline methods.

In general, Figure \ref{fig:scatter} shows that EMOSAM is located on the Pareto front for five datasets and is only dominated by FAHT on the Credit dataset. Specifically, EMOSAM positions towards the bottom left corner in two datasets, namely Law and Census, while occupying a central position in the remaining four. Compared to HT, HAT, and SAM (the green dots), EMOSAM is located on the left (lower discrimination) in five datasets, including Law, Credit, Adult, Bank and Dutch. However, EMOSAM is unbeneficial in model error in comparison with these fairness-unaware methods that only focus on accuracy in their performance. Notably, EMOSAM maintains a competitive position when contrasted with the fairness-aware methods (blue and pink dots). For instance, in the Adult dataset, EMOSAM is located in the middle, while FAHT, FOMOS and FAS Stream tend to be in the lower right corner (higher discrimination and lower error). Moreover, FABBOO is always on the upper left corner with very low discrimination but very high error. This means EMOSAM maintains a better balance between model error and discrimination. Overall, EMOSAM achieves better performance than both fairness-unaware and fairness-aware methods.

\begin{figure*}[!tb]
     \centering
     \subfloat[Model error]{
    \includegraphics[width=0.3\textwidth]{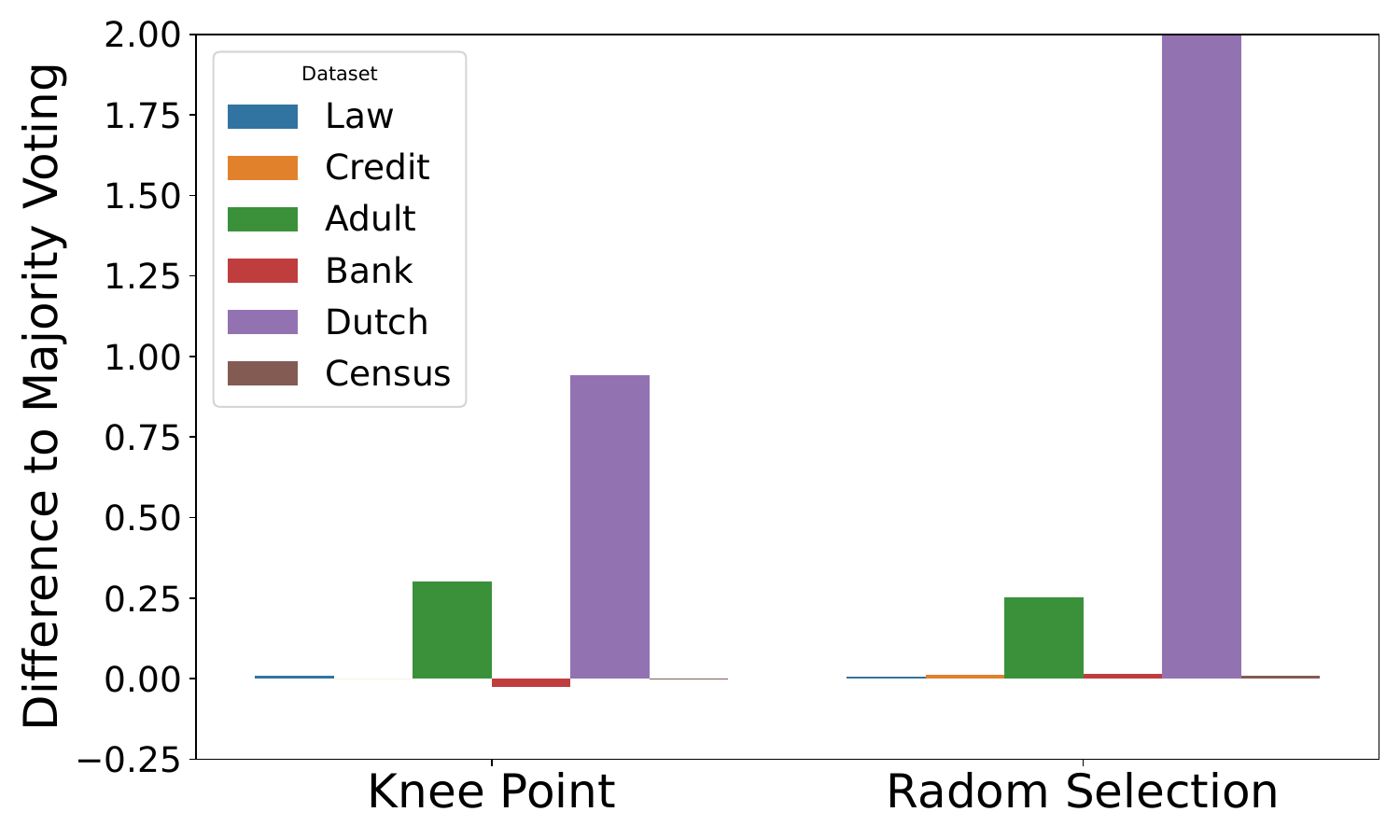}\label{fig:wserror}
    }
    \hfill
     \subfloat[Discrimination]{
    \includegraphics[width=0.3\textwidth]{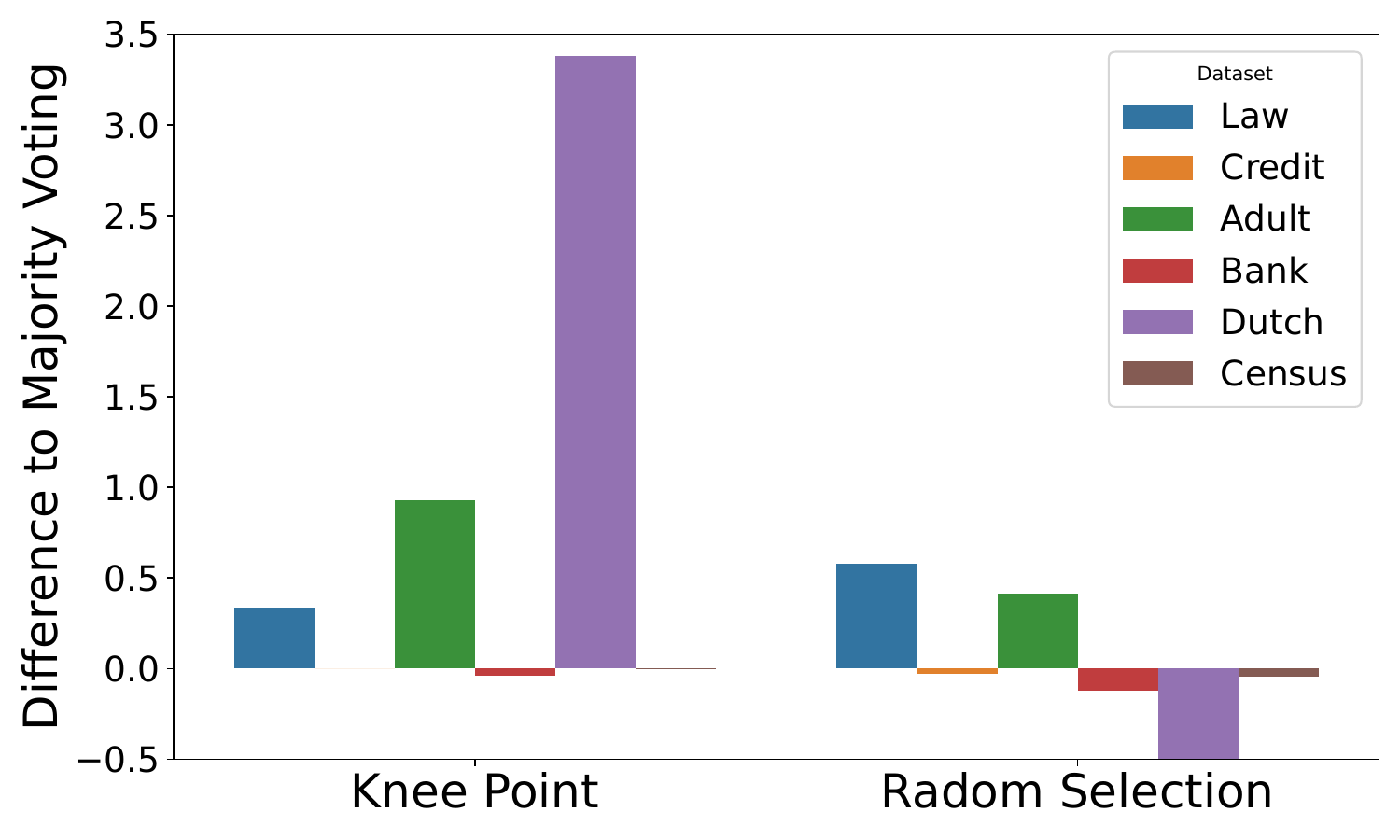}\label{fig:wsdisc}
    }
    \hfill
    \subfloat[Processing Time]{
    \includegraphics[width=0.3\textwidth]{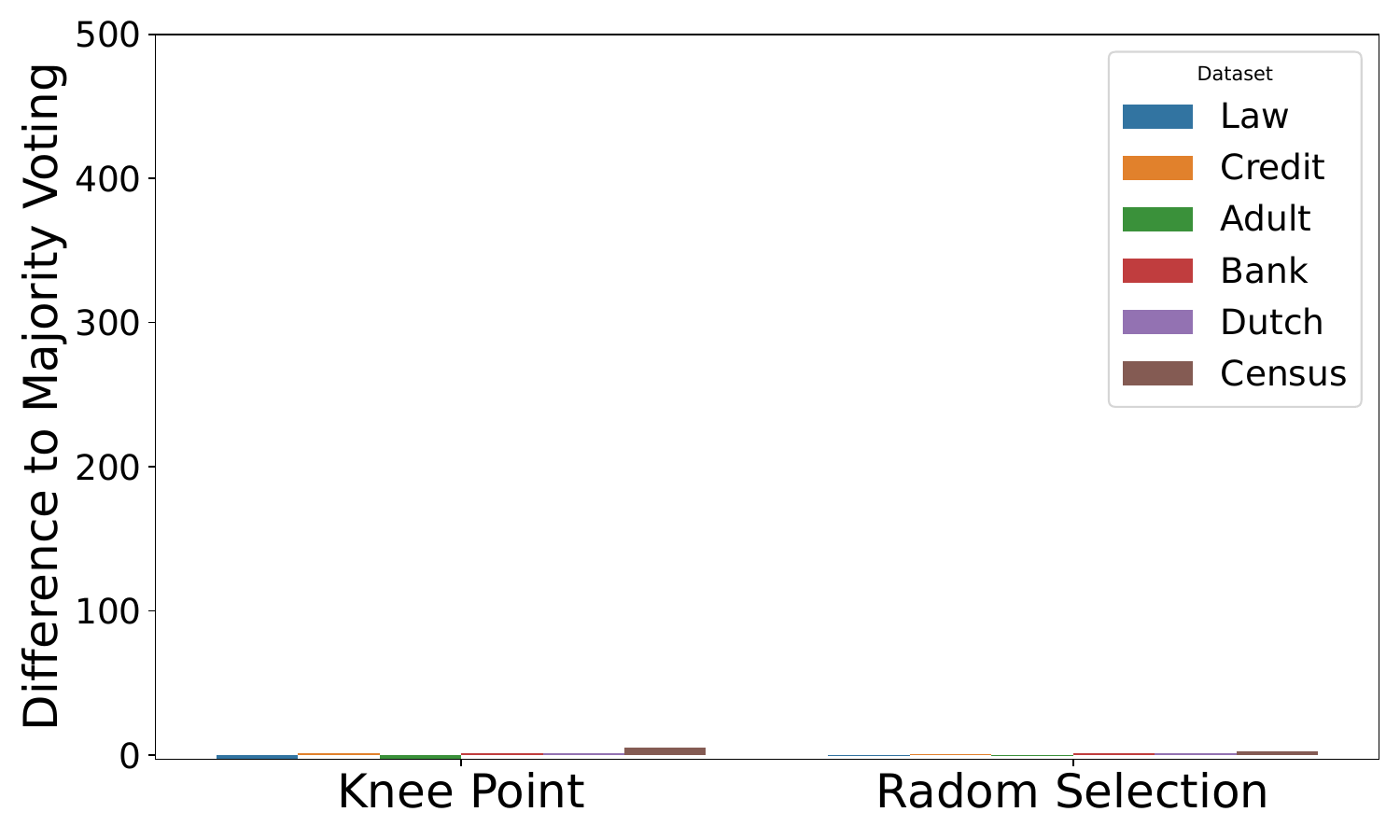}\label{fig:wsruntime}
    }
  \caption{Comparison of weight selection methods. A positive bar means the Majority Voting (used by EMOSAM) is better.}
  \label{fig:leader}
\end{figure*}
\begin{figure*}[!tb]
     \centering
     \subfloat[Model error]{
    \includegraphics[width=0.3\textwidth]{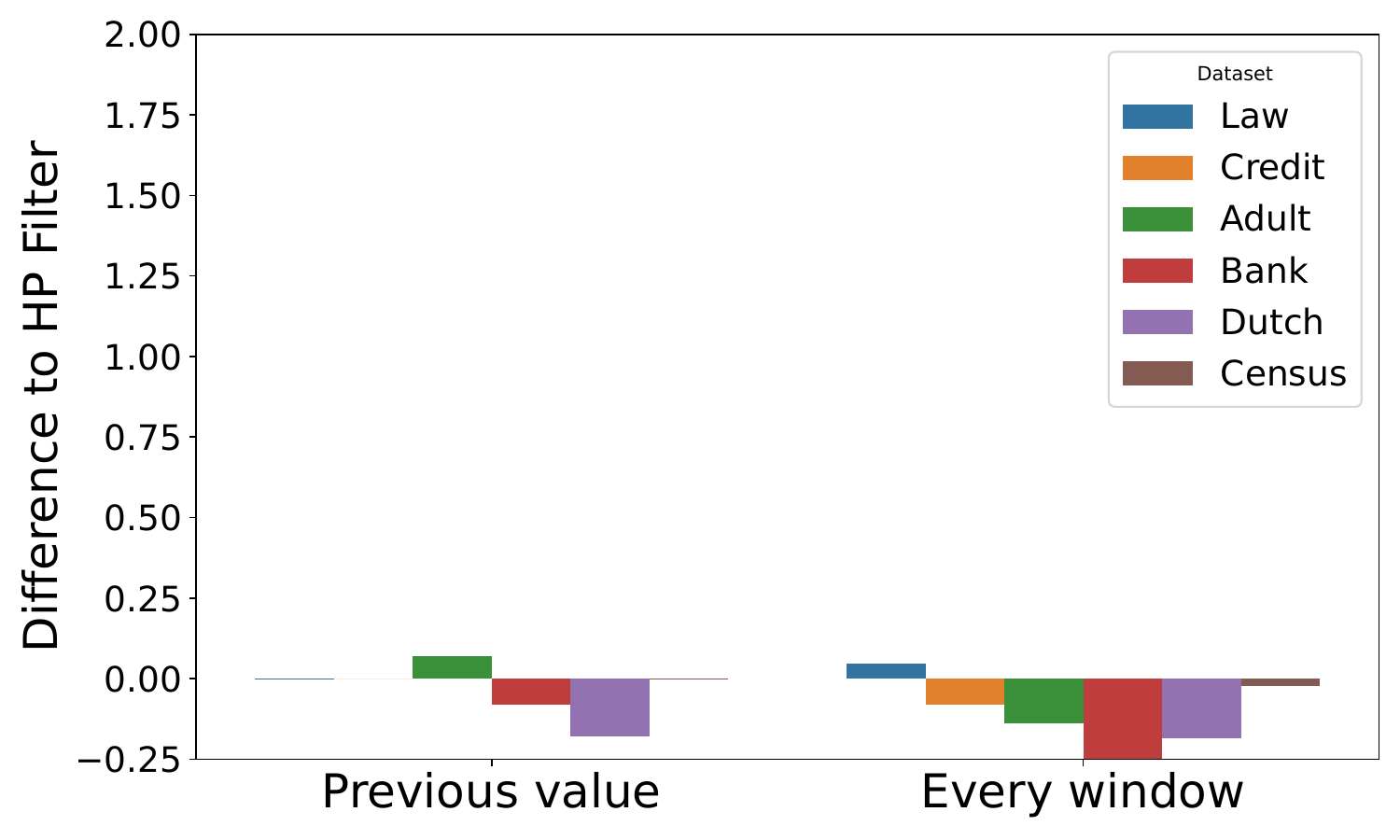}\label{fig:trierror}
    }
    \hfill
     \subfloat[Discrimination]{
    \includegraphics[width=0.3\textwidth]{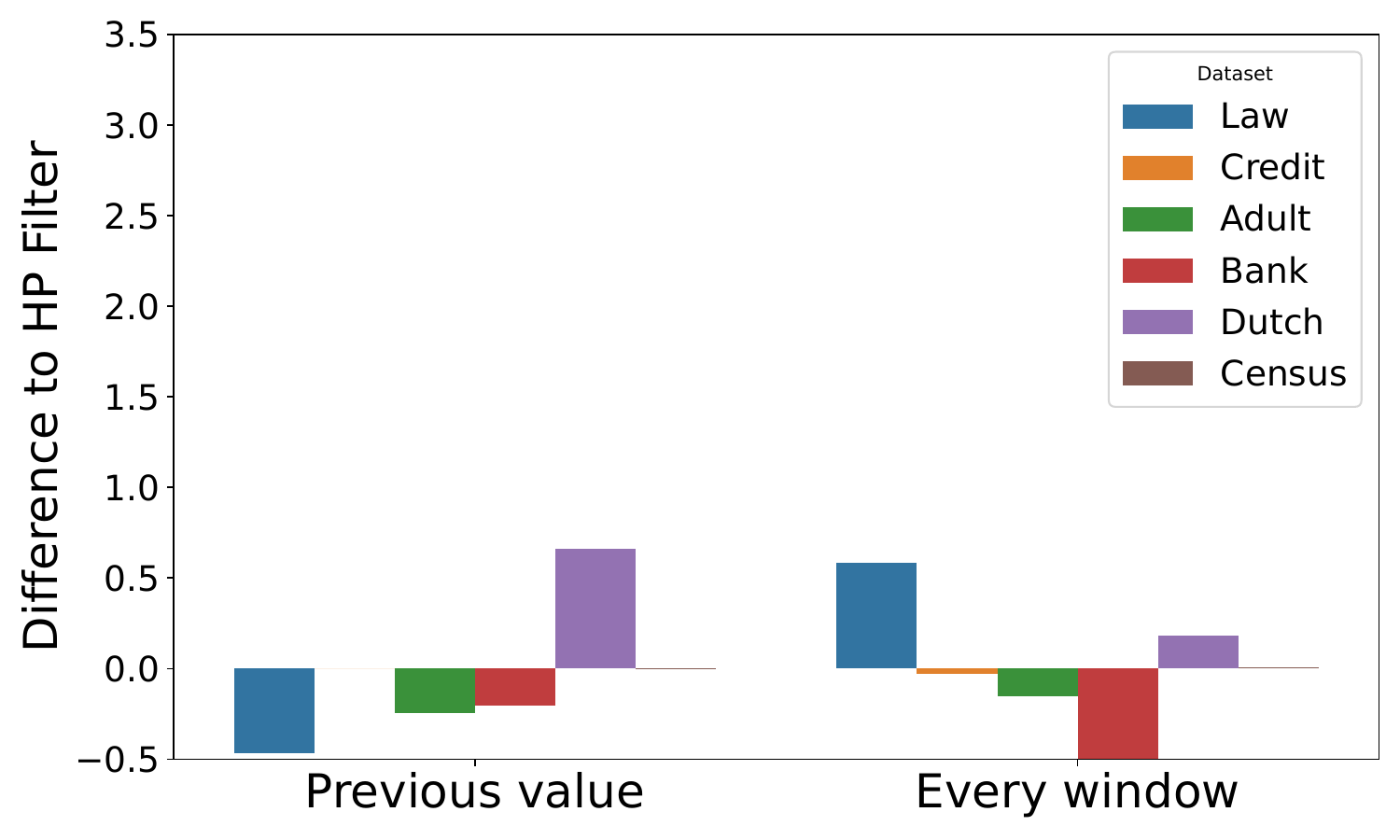}\label{fig:tridisc}
    }
    \hfill
    \subfloat[Processing Time]{
    \includegraphics[width=0.3\textwidth]{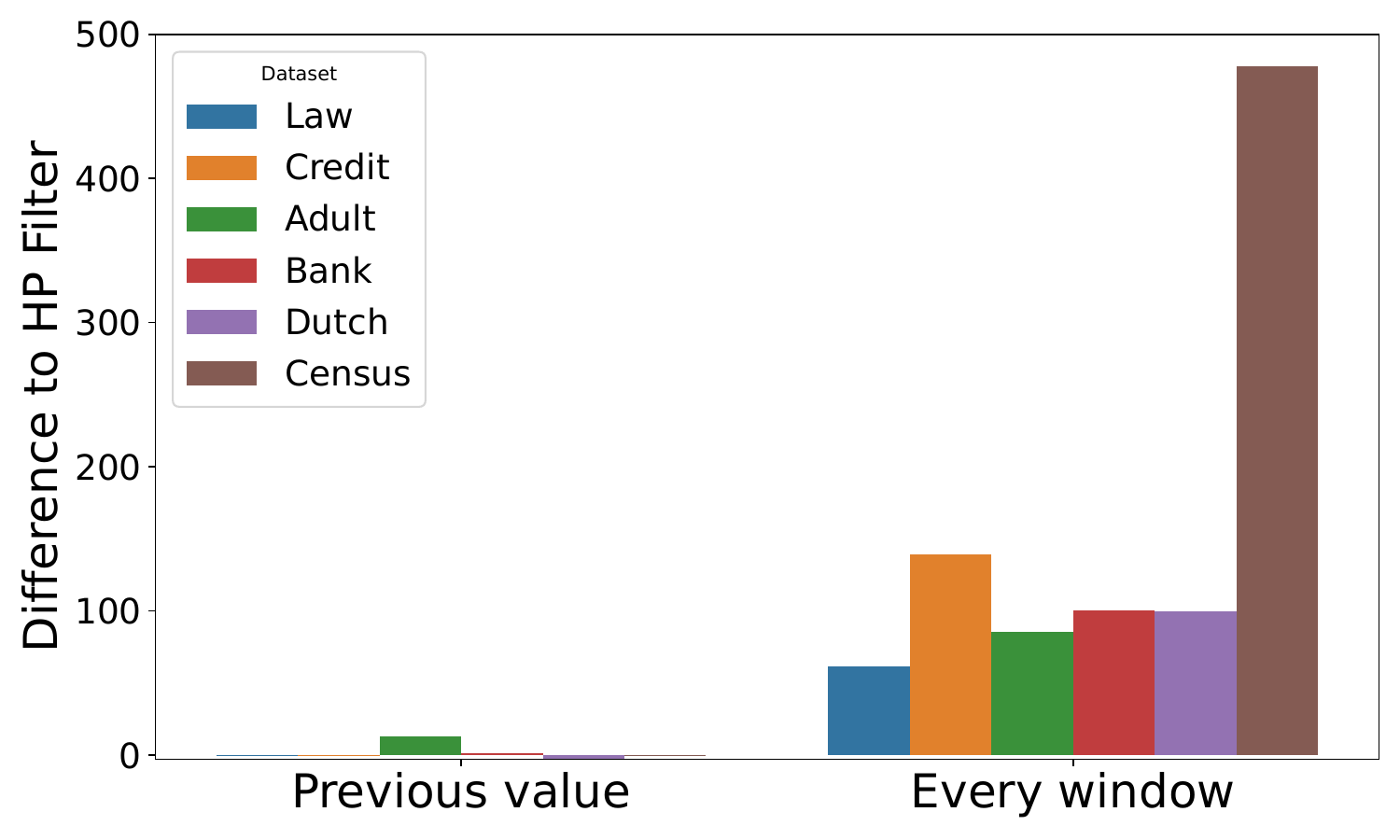}\label{fig:triruntime}
    }
    \caption{Comparison of triggering approaches. A positive bar means the HP filter approach (used by EMOSAM) is better.}
  \label{fig:trigger}
\end{figure*}

\subsection{Ablation Study}
\label{sec:comparewithother}

We further investigate the effectiveness of the two key components of our EMOSAM method: Majority Voting for selecting feature weights and HP filter as a trigger for optimising feature weights. 

\subsubsection*{Feature Weight Selection}

To examine the impact of the Majority Voting method on feature weight selection used by EMOSAM, we replace it with two alternative approaches:
\begin{itemize} 
\item Random Selection, where a feature weight is selected randomly from the Pareto front, 
\item Knee Point, where the Knee point on the Pareto front is selected~\cite{zhang2014knee}. 
\end{itemize}
To ensure a fair comparison, all other settings remain unchanged.

Figure~\ref{fig:leader} shows the difference between the Knee Point / Random Selection approach and the Majority Voting method in terms of model error (\ref{fig:wserror}), discrimination (\ref{fig:wsdisc}), and processing time (\ref{fig:wsruntime}). A positive bar indicates that the Majority Voting method outperforms the other approach, while a negative bar suggests the opposite. In general, the Majority Voting approach achieves lower error and discrimination values while requiring a similar computational time compared to the other approaches. It is worth noting that Figure~\ref{fig:leader} uses the same range as Figure~\ref{fig:trigger}.

In terms of model error, the Majority Voting method consistently achieves the best results when compared to the other two methods. Specifically, according to Wilcoxon signed-rank tests that utilise the mean errors from all six datasets, our Majority Voting method significantly outperforms the Random Selection approach with a p-value of 0.03. Although the errors produced by the Majority Voting and Knee Point methods are not significantly different, our method still manages to generate a slightly lower mean error of 15.53\% compared to 15.81\% from the Knee Point method. Upon closer examination, it is evident that our Majority Voting method performs exceptionally well on the Dutch dataset, reducing the model error by 2\% and 1\% in comparison to the Random Selection and Knee Point approaches, respectively.

Regarding model discrimination, our Majority Voting method consistently generates equally good or better results than the Knee Point method,, especially evident in the Dutch and Adult datasets. In combination with error results, our Majority Voting method dominates the Knee Point method on the Dutch and Adult datasets. The Random Selection method also performs well, with a slightly lower mean discrimination rate of 8.16\%  compared to that of the Majority Voting method (8.25\%). However, according to the Wilcoxon test, there is no statistically significant difference between these results.

The processing time for the three weight selection methods is similar, with a difference of less than 3\%. It is worth noting that the changes in processing time are not as crucial as the changes in model error or discrimination, as a 3\% increase in processing time is manageable. Furthermore, the Wilcoxon tests show no statistically significant difference in the runtimes of the three methods.

\subsubsection*{Weight Optimisation Trigger}
To investigate the effects of the HP filter as a trigger for feature weight optimsation, we substitute it with the following two alternative approaches while keeping the other settings unchanged in EMOSAM: 
\begin{itemize}
    \item Every Window, where feature weight optimsation is triggered in every window, and 
    \item Previous Value, where feature weight optimsation is triggered only if the discrimination in the current window is greater than that of the previous window by a threshold $\theta$, which is set to 7\% in our experiment.
\end{itemize}

In Figure~\ref{fig:trigger}, the HP filter method produces similar outcomes to the Previous Value approach in terms of model error, discrimination, and runtime, as evidenced by the Wilcoxon tests across the six datasets. However, since our HP filter method considers the trend of multiple windows, it is more robust and less sensitive to randomness than the Previous Value approach. Considering a scenario where the model discrimination consistently increases in each window but the difference is below the threshold $\theta$, the Previous Value method will not trigger feature weight optimisation, even when the discrimination reaches 100\%.

In comparison to optimising feature weights in every window, our HP filter method generates slightly larger model errors and discrimination rates but with no statistically significant difference, as confirmed by the Wilcoxon signed rank tests. However, our HP filter method requires significantly less runtime. It is worth noting that the Every Window approach can be considered the ideal scenario for our EMOSAM for producing model errors and discrimination rates. This outcome implies that our HP filter effectively identifies the appropriate windows for optimising feature weights when necessary, while still maintaining a manageable runtime.

\begin{figure}[h]
     \centering
    \subfloat[Discrimination]{
    \includegraphics[width=0.45\textwidth]{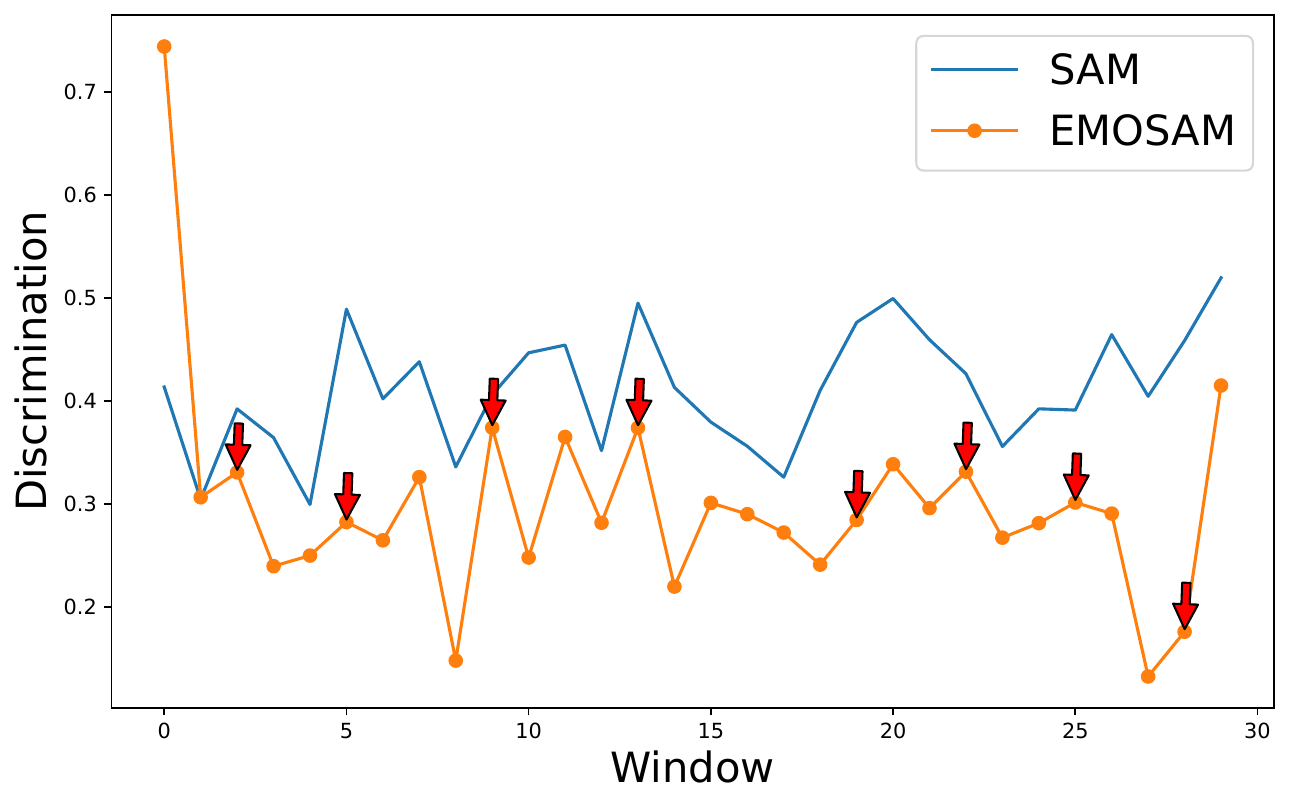}
    }
    \vspace{-3mm}
    \hfill 
     \subfloat[Accuracy]{
    \includegraphics[width=0.45\textwidth]{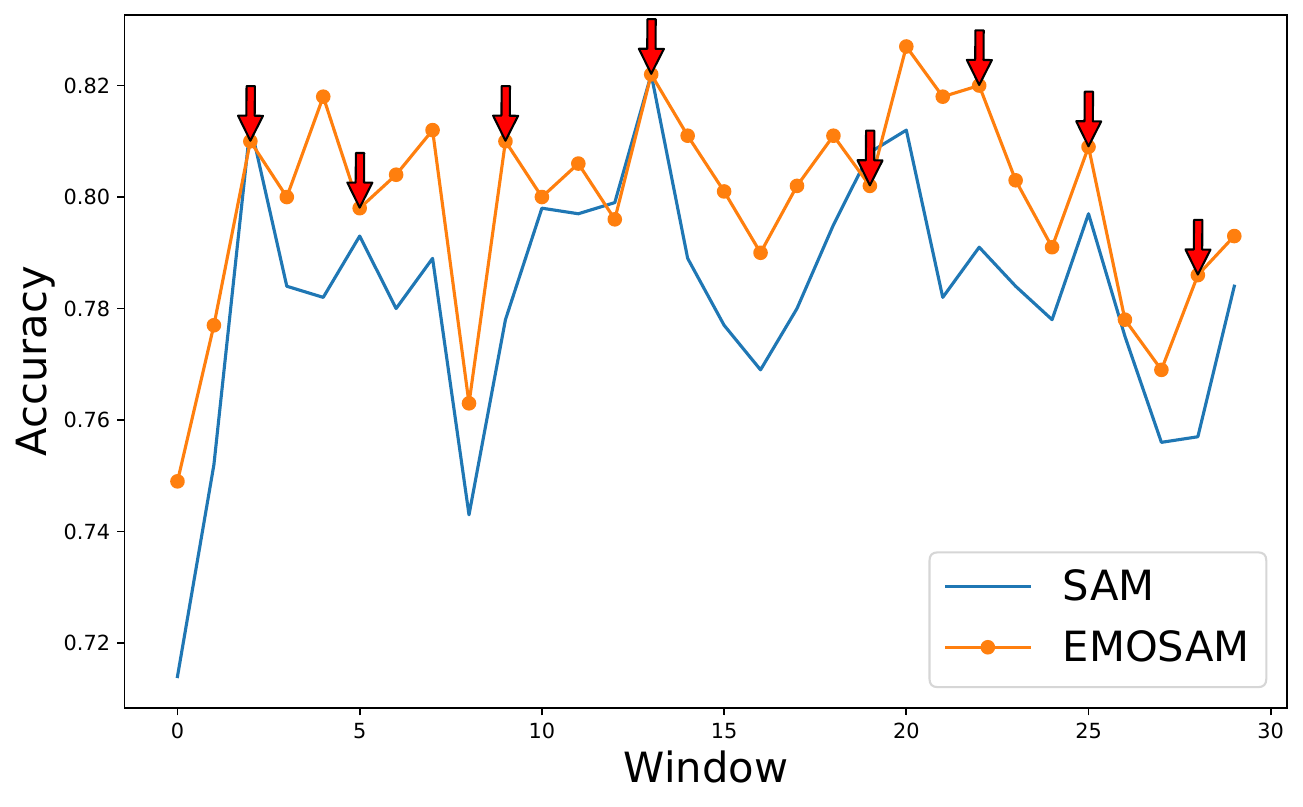}
    }
\vspace{-3mm}
  \caption{The effect of feature weight optimisation on the Dutch dataset over the first 30 windows.}
  \label{fig:dutch}
  \vspace{-6mm}
\end{figure}

To further demonstrate the effectiveness of the HP filter, we present the results of feature weight optimisation on the Dutch dataset in Figure~\ref{fig:dutch}. The figure shows the accuracy and discrimination of EMOSAM and SAM over time, with red arrows indicating when feature weight optimisation is triggered. It can be seen that when the discrimination trend increases, such as in Window 2, the HP filter successfully triggers the optimisation of feature weights to reduce discrimination in the subsequent windows. Interestingly, this also leads to an improvement in model accuracy, as it is one of the optimisation objectives. As a result, EMOSAM achieves a lower discrimination rate and higher accuracy compared to SAM over the 30 windows, and the difference is more significant immediately after applying feature weight optimisation. In the Credit and Census datasets, where the model discrimination is small, there is no need to consider discrimination, and therefore, feature weight optimisation is never triggered, resulting in EMOSAM degenerating to SAM, as shown in Figure~\ref{fig:scatter}. We have also tested multiple values $\{5\%, 10\%, 15\%\}$ for the trend threshold $\Phi$ of the HP filter and observed that it does not significantly change the results. This further demonstrates the robustness of our HP filter method. 

\section{Conclusion}
\label{sec:conclusion}

This study presented an innovative evolutionary multi-objective optimisation (EMO) approach aimed at improving both fairness and accuracy in data stream ML models. We introduced EMOSAM, a novel incremental classification method designed to develop fairness-aware SAMKNN classifiers for stream classification. In EMOSAM, an EMO component optimises feature weights, effectively balancing accuracy and discrimination.

The experiment results on six datasets show that EMOSAM effectively and efficiently balances the conflicting objectives to dominate the baseline methods in many cases. The ablation study further validated our proposed strategies for EMO triggering and the selection of non-dominated solutions. The activation of the EMO component can consistently enhance the classification performance of SAMKNN over time, while the majority voting strategy proposed in this paper can significantly improve both the accuracy and fairness of the classification model. 

Moving forward, our research will focus on several key areas. We plan to investigate advanced EMO algorithms for a better balance between fairness and accuracy, with an emphasis on adapting to dynamic data streams. Additionally, we will explore the scalability of EMOSAM across larger and more varied datasets to assess its robustness in different scenarios. Integrating explainability into EMOSAM will also be a priority, enhancing model transparency and understanding of fairness and accuracy determinants. Lastly, we aim to include more comprehensive fairness criteria and metrics, broadening our approach to fairness in machine learning. These steps are crucial for advancing fair and accurate machine learning, especially in the complex world of data streams.

\bibliography{reference}
\end{document}